\documentclass[preprint,12pt]{elsarticle}

\usepackage{amssymb}

\usepackage{amsmath,amssymb,amsfonts}
\usepackage{graphicx}
\usepackage{siunitx}
\usepackage{xcolor}
\usepackage{booktabs}
\usepackage{makecell}
\usepackage{multirow}
\usepackage{arydshln}
\usepackage{enumitem}
\usepackage{subcaption}
\usepackage{caption}

\journal{Applied Soft Computing}

\begin{document}

\begin{frontmatter}

\title{NeuroVascU-Net: A Unified Multi-Scale and Cross-Domain Adaptive Feature Fusion U-Net for Precise 3D Segmentation of Brain Vessels in Contrast-Enhanced T1 MRI}

\author[1]{Mohammad Jafari Vayeghan}
\ead{Mhmmd.jafari@ut.ac.ir}

\author[1,2]{Niloufar Delfan}
\ead{ndelfan@yorku.ca}

\author[1]{Mehdi Tale Masouleh}
\ead{m.t.masouleh@ut.ac.ir}

\author[3]{Mansour Parvaresh Rizi}

\ead{parvareshrizi.m@iums.ac.ir}

\author[1,4]{Behzad Moshiri}

\ead{moshiri@ut.ac.ir}

\cortext[1]{Behzad Moshiri, Mansour Parvaresh Rizi}

\affiliation[1]{
  organization={School of Electrical and Computer Engineering, College of Engineering, University of Tehran},
  city={Tehran},
  country={Iran}
}

\affiliation[2]{
  organization={Department of EECS, Lassonde School of Engineering, York University},
  city={Toronto},
  country={Canada}
}

\affiliation[3]{
  organization={Department of Neurosurgery, School of Medicine, Iran University of Medical Sciences},
  city={Tehran},
  country={Iran}
}

\affiliation[4]{
  organization={Department of Electrical and Computer Engineering, University of Waterloo},
  city={Waterloo},
  country={Canada}
}

\begin{abstract}
Precise 3D segmentation of cerebral vasculature from T1-weighted contrast-enhanced (T1CE) Magnetic Resonance Imaging (MRI) is critical for safe and effective neurosurgical planning. However, manual delineation is labor-intensive and prone to inter-observer variability, while existing automated methods often face a trade-off between segmentation accuracy and computational efficiency, hindering clinical adoption. To the best of our knowledge, this study presents NeuroVascU-Net, the first deep learning architecture specifically designed to segment cerebrovascular structures directly from clinically standard T1CE MRI in neuro-oncology patients, addressing a significant gap in the literature that predominantly focuses on Time-of-Flight Magnetic Resonance Angiography (TOF-MRA). NeuroVascU-Net is built upon a dilated U-Net backbone and uniquely integrates two specialized modules: a Multi-Scale Contextual Feature Fusion ($MSC^2F$) module at the bottleneck and a Cross-Domain Adaptive Feature Fusion ($CDA^2F$) module at deeper hierarchical levels. The $MSC^2F$ module captures local and global contextual information through multi-scale dilated convolutions, while the $CDA^2F$ module dynamically integrates domain-specific features, collectively enhancing feature representation while minimizing computational overhead. The model was trained and validated on a uniquely curated dataset of T1CE MRI scans from 137 patients undergoing brain tumor biopsy, with expert annotations by a board-certified functional neurosurgeon. NeuroVascU-Net achieved state-of-the-art performance, demonstrating a Dice Similarity Coefficient (DSC) of 0.8609 and precision of 0.8841, accurately delineating both large and fine vascular structures. Notably, it operates with only 12.4 million parameters, significantly fewer than transformer-based models like Swin U-NetR (15.7 million), ensuring reduced computational load without compromising accuracy. This combination of high precision and efficiency positions NeuroVascU-Net as a powerful, practical solution for computer-assisted neurosurgical planning, with considerable potential to enhance procedural safety and improve patient outcomes.
\end{abstract}

\begin{keyword}
 Deep Learning\sep Contrast Enhanced Magnetic Resonance Imaging\sep Brain Vessel \sep Neuro-navigation\sep Segmentation\sep
\end{keyword}

\end{frontmatter}

\section{Introduction}
Accurate delineation of complex brain anatomical structures, particularly vascular networks and neoplastic lesions, is essential for ensuring the safety and effectiveness of modern neurosurgical procedures. This is especially crucial in guiding interventions such as stereotactic needle biopsies of deep-seated or multifocal intracranial lesions \cite{akshulakov2019current}, where the risk of iatrogenic vascular injuries during needle placement or tissue sampling remains significant, even with minimally invasive techniques \cite{fruhwirth1997vascular,bex2022advances}. Consequently, precise pre-operative identification of both tumors and surrounding vasculature is indispensable, not only for improving diagnostic accuracy but also for enabling detailed surgical planning for procedures like stereotactic biopsies and complex tumor resections \cite{vadhavekar2024advancements, wang2014critical}.

Contrast-enhanced T1-weighted (T1CE) Magnetic Resonance Imaging (MRI) has become a cornerstone imaging modality in this context. Its ability to clearly delineate tumor margins and visualize the intricate cerebrovascular network makes it invaluable for constructing detailed anatomical maps to minimize contact with critical vessels during surgery\cite{shukla2017advanced, young2007advanced}. However, translating this rich imaging data into actionable surgical plans remains challenging due to the conventional reliance on manual segmentation. Manual segmentation is notoriously labor-intensive, time-consuming, and prone to inter-observer variability, which can lead to inconsistencies in surgical planning and negatively impact patient outcomes \cite{despotovic2015mri}.

Automated segmentation, driven by advancements in computational methods like deep learning (DL), offers a transformative solution to these challenges. DL-based approaches enable objective, efficient, and reproducible delineation of anatomical structures \cite{liu2021review}. Precise vascular mapping is particularly critical in tumor resections and biopsies, as it minimizes the risk of inadvertent vascular injury, thereby improving patient safety and surgical outcomes. Additionally, the time-saving benefits of automation allow for rapid data processing, near real-time intraoperative feedback, and seamless integration with robotic-assisted platforms \cite{reddy2023advancements}. These advances reduce the interpretive burden for neurosurgeons, minimize human error, and standardize the identification of critical structures, ultimately enabling safer surgical pathways \cite{monfaredi2024automatic}. 

The segmentation of vascular structures, particularly using Time-Of-Flight Magnetic Resonance Angiography (TOF-MRA), has been extensively explored in the literature, employing various algorithmic approaches \cite{moccia2018blood, goni2022brain}. Early methods primarily relied on rule-based techniques, which varied in classification principles, vessel extraction processes, and levels of manual interaction. These approaches can be categorized into Hessian matrix-based methods \cite{Frangi1998,Sato1998,xiao2012multiscale}, marching filtering \cite{qian2009non}, mathematical morphology \cite{zana2001segmentation,babin2012generalized,dufour2013filtering}, minimal path \cite{forkert20133d,mohan2010tubular,cetin2012vessel}, active contour \cite{li2011level,lorigo2001curves,manniesing2006vessel,manniesing2007vessel,cheng2015accurate}, and graph-based methods \cite{bauer2010segmentation,esneault2009liver}. While foundational, these methods often struggled with the challenges posed by complex vascular geometries, variations in vessel size and intensity, and susceptibility to noise and imaging artifacts \cite{zhao2019segmentation}. 

The introduction of DL, particularly Convolutional Neural Networks (CNNs), marked a paradigm shift in vascular segmentation. Several 2D and 3D CNN-based models have been developed, addressing challenges such as data scarcity, false positives, and computational complexity. For example, Phellan et al. \cite{phellan2017vascular} demonstrated the use of a simple 2D CNN, while subsequent studies incorporated advanced techniques such as Maximum Intensity Projection (MIP) integration \cite{nakao2018deep}, hierarchical labeling \cite{zhao2018semi}, and anatomical region-based data stratification \cite{kandil2018using} to improve accuracy. Hybrid approaches, including the combination of Hidden Markov Random Fields (HMRFs) with DL \cite{fan2020unsupervised} and architectural innovations like DeepVesselNet with 2D cross-hair filters \cite{tetteh2020deepvesselnet}, further enhanced segmentation capabilities. Advanced models such as DeepMedic \cite{kamnitsas2017efficient} and DD-CNN \cite{zhang2020cerebrovascular} introduced dual-pathway designs, dense connections, and Conditional Random Field (CRF)-based post-processing, achieving Dice Similarity Coefficient (DSC) scores as high as 0.97.

The U-Net architecture \cite{ronneberger2015u}, characterized by its encoder-decoder structure and effective skip connections, quickly became a standard in segmentation tasks. Livne et al. \cite{livne2019u} demonstrated the efficacy of a compact 2D U-Net in vessel segmentation, achieving a DSC of 0.88 while maintaining reduced model complexity. Vos et al. \cite{Unet3Dvessel} later extended this approach to 3D patch-based U-Net models, further improving DSC scores (0.72–0.83) on TOF-MRA datasets through data augmentations such as Gaussian blur, rotation, and flipping.

Subsequent research refined and enhanced the U-Net framework. BraveNet, introduced by Hilbert et al. \cite{Bravenet}, integrated multiscale context aggregation and deep supervision into a 3D U-Net architecture, achieving higher performance with a DSC of 0.93. Attention mechanisms, which allow models to focus on relevant image features, have also gained significant traction. For instance, Abbas et al. \cite{attentionunet} incorporated attention U-Net designs for improved segmentation of TOF-MRA volumes, reporting a DSC of 0.91 on the TTKU-L dataset. Min et al. \cite{MSFE} introduced a U-Net variant equipped with Channel and Spatial Attention Modules (CABM) and Multi-Scale Feature Extraction (MSFE), achieving a DSC of 0.882. The application of dilated convolutions and residual connections, as explored by Amer et al. \cite{mdaunet} for lung CT segmentation, further underscores their potential for cerebrovascular segmentation.

Despite significant advancements in vascular segmentation using TOF-MRA and CTA, there remains a notable gap in research on the automated segmentation of cerebrovascular structures directly from T1CE MRI scans. T1CE MRI provides unique advantages, such as simultaneous visualization of tumors, surrounding edema, and contrast-enhancing vessels (both arteries and veins), which is critical for comprehensive neurosurgical planning \cite{lee2005preoperative,delfan2025advancing}. However, it also presents challenges, including variable contrast uptake, leakage into tumor tissue that can obscure vessel-tumor boundaries, and a generally more complex visual appearance compared to flow-based angiographic sequences. As a result, the development and validation of DL models for robust cerebrovascular segmentation in the challenging context of tumor-bearing brains remain largely unexplored \cite{abidin2024recent}.

This study addresses this unmet need by introducing a novel 3D U-Net-based architecture, NeuroVascU-Net, tailored for the segmentation of cerebrovascular structures from T1CE MRI in patients with brain tumors. Key contributions include the creation of a dedicated T1CE MRI dataset of 137 patients with expert annotations and the development of the NeuroVascU-Net model, which incorporates specialized blocks to capture both local and global contextual information. This paper details the dataset, architecture, training methodology, segmentation performance, and discusses its potential clinical applications in neurosurgical planning.

\section{Methods and Materials}
This section presents the systematic methodology and resources employed in this study to achieve accurate cerebrovascular segmentation from T1CE MRI scans. It details the dataset acquisition and annotation process, pre-processing pipeline, model architecture design, evaluation metrics, and optimization strategies. 
\subsection{Data Collection and Annotation}
\begin{figure}
  \centering 
  
  \begin{subfigure}[b]{0.35\textwidth} 
    \centering
    \includegraphics[width=\linewidth]{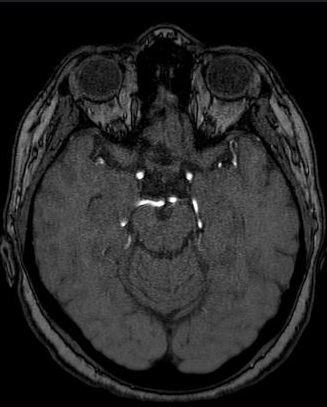}
    \caption{} 
    \label{fig:tof}
  \end{subfigure}%
  \hspace{0.5em} 
  \begin{subfigure}[b]{0.37\textwidth}
    \centering
    \includegraphics[width=\linewidth]{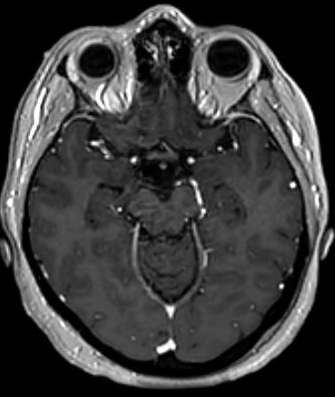}
    \caption{}
    \label{fig:t1ce}
  \end{subfigure}
  \caption{Comparison of vascular and tissue visualization in TOF-MRA and T1CE MRI. (a) TOF-MRA highlights large, fast-flowing arteries with limited soft tissue contrast and poor visibility of small or slow-flow vessels. (b) T1CE MRI provides enhanced anatomical detail and visualizes both arteries and veins, including smaller vessels near pathological tissue, but introduces complexity in vessel segmentation due to nonspecific enhancement.} 
  \label{fig:Modality}
\end{figure}

The selection of imaging modality significantly impacts the visualization of cerebrovascular structures, which is crucial for pre-operative neurosurgical planning. TOF-MRA relies on flow-related enhancement to effectively highlight large, fast-flowing arteries. However, as shown in Fig.~\ref{fig:Modality} (a), TOF-MRA is limited in soft tissue contrast and often fails to adequately visualize small, slow-flowing vessels and venous structures. While its focused arterial visualization simplifies segmentation, it frequently results in incomplete vascular maps, particularly around tumor margins or in anatomically complex regions.

In contrast, T1CE MRI uses a gadolinium-based contrast agent to enhance areas of increased perfusion or disrupted blood-brain barriers, typically associated with neoplastic tissue. As depicted in Fig.~\ref{fig:Modality}(b), T1CE MRI provides superior soft-tissue contrast and enables visualization of both arteries and veins, including smaller vessels that are often missed by TOF-MRA. This detailed anatomical information is critical for surgical planning, particularly in understanding the tumor-vasculature interface.

Despite these advantages, T1CE imaging poses unique challenges for vessel segmentation. Its non-specific enhancement can cause hyperintense signals in tumors, dura, and inflamed tissues, leading to heterogeneous vessel intensities and signal overlap with surrounding structures. This complexity makes manual segmentation a highly time-consuming and expertise-dependent process, while conventional automated methods struggle to accurately distinguish vessels from other enhancing regions. These limitations are especially problematic for identifying fine or subtly enhancing vessels that are critical for safe surgical procedures.

To address these challenges, this study utilized a prospectively collected dataset comprising 1.5 Tesla (T) T1CE MRI scans with approximately 1-mm in-plane resolution from 137 patients undergoing stereotactic brain biopsies at Rasoul Akram Hospital, Tehran, Iran, between 2023 and 2024. Ethical approval was obtained from the Iran University of Medical Sciences (Approval No: IR.IUMS.FMD.REC.1403.377). Each MRI scan contains approximately 160 axial slices with a resolution of 256×256 pixels. All scans were annotated meticulously by a board-certified functional neurosurgeon (M.PR) with over 20 years of experience, using the 3D Slicer software \cite{slicer}.

\subsection{Data Pre-processing}

A comprehensive pre-processing pipeline was implemented to optimize the T1CE MRI data for deep learning-based segmentation. The pipeline was designed to standardize input data, minimize artifacts, reduce computational complexity, and improve model generalization. The first step involved skull stripping to remove non-brain tissues from the T1CE MRI scans. This was performed using the HD-BET algorithm \cite{isensee2019automated}. After skull stripping, N4 bias field correction was applied to address intensity non-uniformities caused by magnetic field inhomogeneities and scanner imperfections. This correction, implemented using the SimpleITK library, reduces low-frequency artifacts that can obscure anatomical details, ensuring more consistent intensity values across the image. By mitigating these artifacts, the N4 correction step enhances the accuracy of downstream segmentation tasks, particularly in regions with variable intensity values due to contrast enhancement.

Following bias field correction, voxel intensities were normalized to a standardized range between 0 and 255 to ensure uniformity across all samples. The MRI volumes were then resized to a consistent spatial dimension of 192 × 192 × 128 voxels. This resizing process involved cropping or padding each volume, focusing the model’s learning on relevant anatomical structures while removing uninformative background regions. These steps ensured that the dataset was optimized for training by reducing variability and highlighting essential features.

To further improve model robustness and reduce the risk of overfitting, several data augmentation techniques were applied. These included random flipping along the y-axis with a  30\% probability to introduce spatial variability and Gaussian noise augmentation to increase the diversity of the training data. Specifically, Gaussian noise with a mean of 0 and a standard deviation of 0.01 was applied exclusively to background regions. This targeted approach preserved critical foreground features while simulating variability in uninformative areas of the image.

\begin{figure*}
  \centering
  \includegraphics[width=1.0\textwidth]{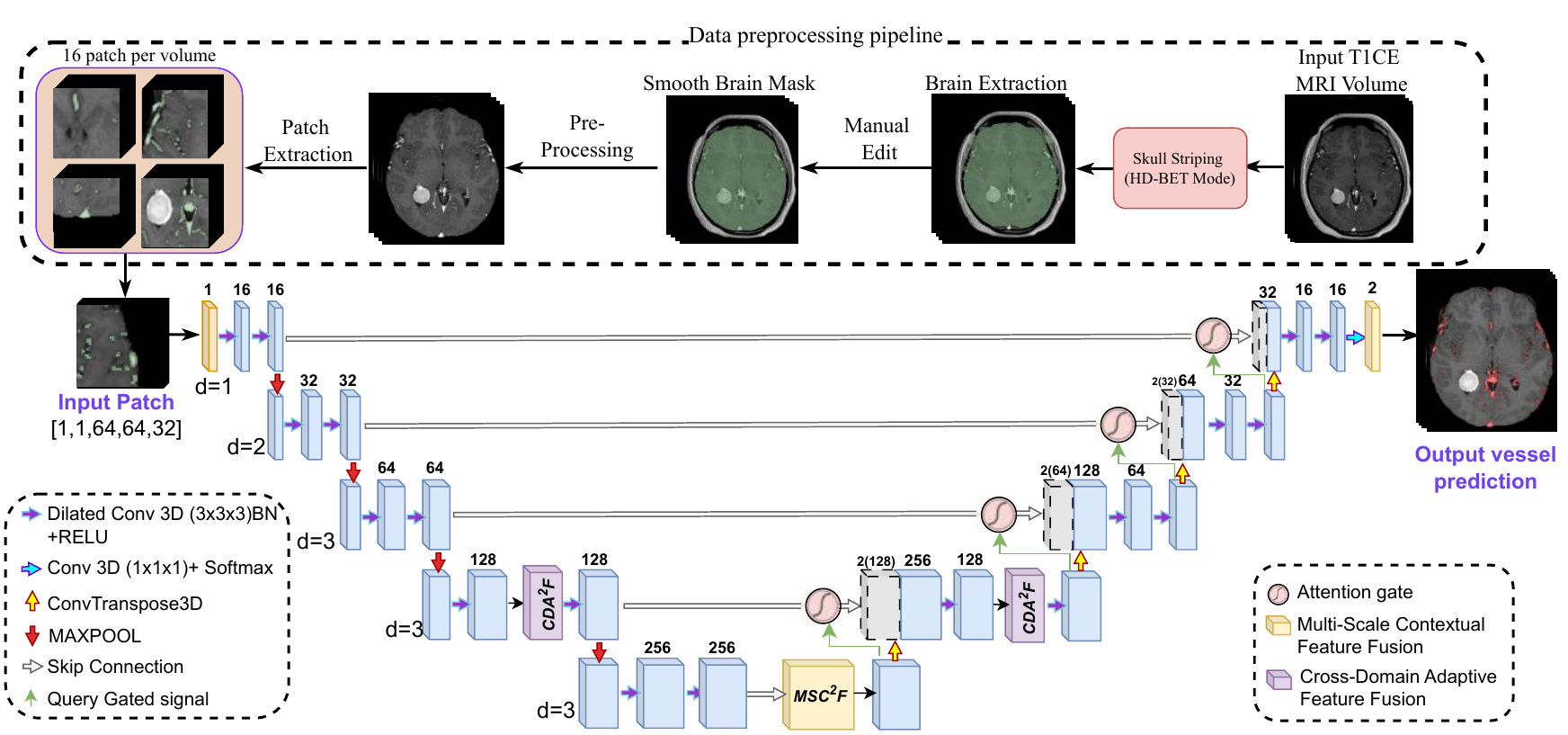}
  \caption{Segmentation process and proposed NeuroVascU-Net network structure.}
  \label{fig:FUSE_architecture}
\end{figure*}

\subsection{Model Architecture}

The Spatial-Enriched Fusion Attention-based U-Net (NeuroVascU-Net), illustrated in Fig~\ref{fig:FUSE_architecture},  was developed to accurately segment 3D cerebrovascular structures from T1CE MRI scans. The architecture is grounded in the widely-used 3D U-Net encoder-decoder framework \cite{Unet3Dvessel}, which is known for its effectiveness in medical image segmentation tasks. To address the complex nature of cerebrovascular segmentation, the NeuroVascU-Net introduces several novel components, including dilated convolutions, attention-gated skip connections, and the integration of two specialized modules: the  Cross-Domain Adaptive Feature Fusion ($CDA^2F$) and the Multi-Scale Contextual Feature Fusion ($MSC^2F$) module. Together, these innovations allow the model to capture hierarchical, multi-scale, and anatomically relevant features, enabling precise segmentation in challenging imaging environments.

At its foundation, the NeuroVascU-Net replaces standard convolutional layers with dilated convolutions across all hierarchical levels, which systematically expand the receptive field and enable the model to capture broader contextual information without increasing computational cost. This design choice ensures that the model can effectively learn the intricate patterns of cerebrovascular structures, including varying vessel sizes and shapes, while remaining computationally efficient.

The encoder path consists of five hierarchical levels with feature-map dimensionalities \(C=\{16,\,32,\,64,\,128,\,256\}\). The first three levels employ sequential blocks of dilated convolutions, each followed by batch normalization and ReLU activation. At each level, feature maps are downsampled by a factor of two using 3D max-pooling, progressively capturing higher-level spatial features. At the fourth level, the model incorporates the $CDA^2F$ module, a specialized feature extraction block designed to learn abstract and geometrically invariant representations. This module fuses features from four parallel branches: 3D Involution, Frequency-Spatial Attention, Spherical CNNs, and a 3D Depthwise ConvNeXt block. Each branch is tailored to address specific challenges in cerebrovascular segmentation, such as spatial variability, frequency-domain enhancement, rotational equivariance, and multi-scale contextual learning. The outputs of these branches are combined via summation and residual connections, creating a rich feature representation that is robust to anatomical variability and capable of capturing fine vascular details.

At the network’s deepest layer, the bottleneck is formed by the $MSC^2F$ module, which is designed to fuse multi-scale spatial context, edge-boundary details, and frequency-domain information into a cohesive feature digest. The $MSC^2F$ module employs Atrous Spatial Pyramid Pooling (ASPP) with anisotropic dilation rates to capture features at multiple scales, allowing the network to discern both large and fine vascular structures. Additionally, the module uses a Laplacian of Gaussian (LoG) filter to enhance vessel boundaries by detecting sharp intensity transitions and tube-like structures, which are characteristic of cerebrovascular anatomy. A Frequency-Spatial Attention mechanism further refines features by amplifying vessel-specific frequency components and suppressing irrelevant signals in the background. The integration of edge and frequency tokens with the skip connection features ensures that the decoder receives enriched and highly refined feature maps, enabling accurate segmentation of both large and small vessels.

Information flows from the encoder to the decoder via attention-gated skip connections \cite{attentionunet}. These skip connections use a grid attention mechanism to suppress irrelevant background signals and amplify vessel-specific features before concatenation with the upsampled decoder features. This ensures that only the most relevant information is passed to the decoder, improving segmentation accuracy and reducing noise.

The decoder path mirrors the encoder structure, progressively reconstructing the segmentation map with high fidelity. Following upsampling through transposed convolution, feature maps are concatenated with attention-gated skip connection features, which have been refined to enhance vessel-specific information. At the fourth decoder level, the $CDA^2F$ module is reintroduced to further process the newly fused features, ensuring consistent refinement during upsampling. The shallower decoder levels rely on blocks of dilated convolutions to iteratively reconstruct the segmentation map, with the final output generated by a \(1 \times 1 \times 1\) convolutional layer. This layer performs voxel-wise predictions, producing the final segmentation map with high spatial resolution and anatomical accuracy. 

\subsubsection{$MSC^2F$ Module}
\begin{figure}
  \centering 
  \begin{subfigure}[b]{0.49\linewidth} 
    \centering
    \includegraphics[width=\linewidth]{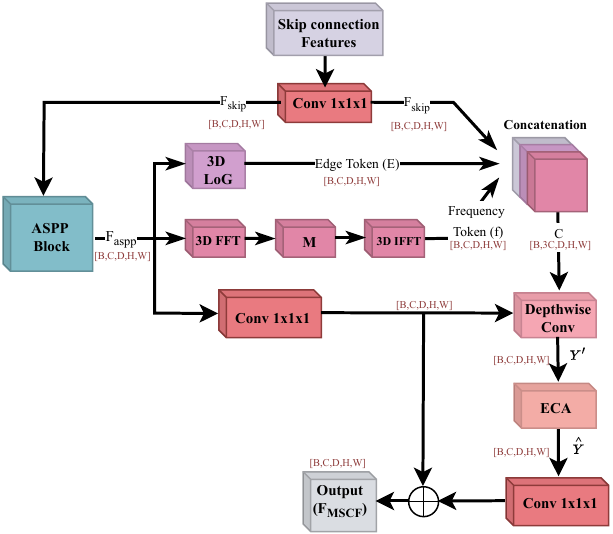}
    \caption{} 
    \label{fig:tof}
  \end{subfigure}
  \begin{subfigure}[b]{0.49\linewidth}
    \centering
    \includegraphics[width=\linewidth]{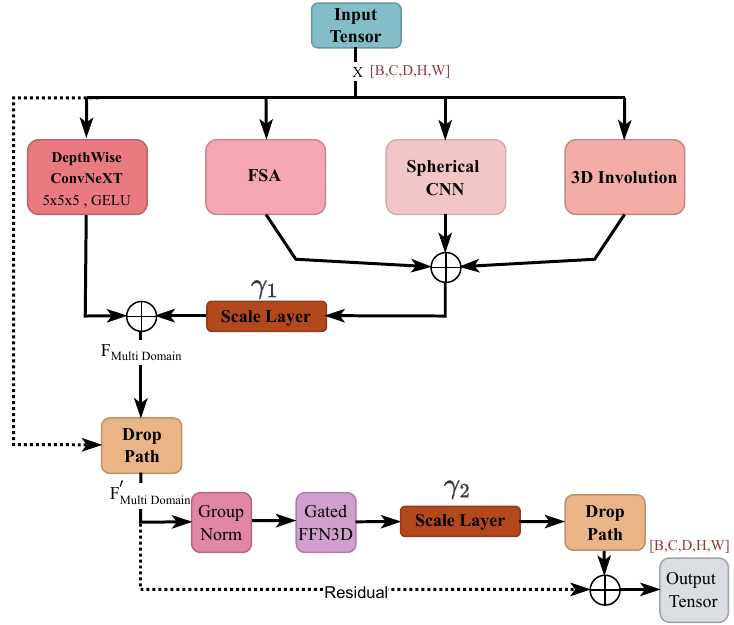}
    \caption{}
    \label{fig:t1ce}
  \end{subfigure}
  
  \caption{Detailed structure blocks for NeuroVascU-Net: (a) $MSC^2F$ block; (b) $CDA^2F$ block.}
  \label{fig:sefa-blocks}
\end{figure}

The $MSC^2F$ module, Fig.~\ref{fig:sefa-blocks}(a), constitutes the bottleneck of the NeuroVascU-Net architecture and is strategically designed to refine and fuse multi-scale spatial, edge-boundary, and frequency-domain features for enhanced cerebrovascular segmentation. This module is critical for creating a comprehensive feature representation that captures both large-scale contextual information and fine-grained vessel details.

The $MSC^2F$ module begins by processing the input skip connection features \(\mathbf{F}_{\mathrm{encoder}} \in \mathbb{R}^{B \times C \times D \times H \times W}\) using an Atrous Spatial Pyramid Pooling (ASPP) block \cite{ASPPantro}. ASPP employs an anisotropic approach to dilation rates \(\{(1,1,1),(1,2,2),(1,3,3)\}\), applied uniformly across all axes. This method enables the module to capture contextual information at multiple spatial scales simultaneously, allowing the network to identify both large vessels and smaller, intricate vascular structures. To enhance sensitivity to anatomical boundaries, the multi-scale feature maps \(\mathbf{F}_{\mathrm{ASPP}} \in \mathbb{R}^{B \times C \times D \times H \times W}\) are convolved with a 3D Laplacian of Gaussian (LoG) filter \cite{Laplacianguassian}, which generates an edge-enhanced feature map \(\mathbf{E}\) by highlighting sharp intensity transitions and tubular structures characteristic of vascular anatomy. The mathematical formulation of the LoG filter is defined as:

\begin{equation}
k_{\mathrm{LoG}}(\mathbf{r};\sigma) = -\frac{1}{\pi\sigma^{4}} \left(1 - \frac{\|\mathbf{r}\|^{2}}{2\sigma^{2}}\right) \exp\left(-\frac{\|\mathbf{r}\|^{2}}{2\sigma^{2}}\right),
\end{equation}

where \(\mathbf{r} = (x, y, z)\) represented spatial coordinates and \(\sigma\) denoted the standard deviation of the Gaussian. The convolution of this kernel with \(\mathbf{F}_{\mathrm{ASPP}}\) produces the edge-enhanced feature map:

\begin{equation}
\mathbf{E} = k_{\mathrm{LoG}} * \mathbf{F}_{\mathrm{ASPP}}.
\end{equation}

In parallel, a Frequency-Spatial Attention (FSA) module processes the same ASPP features by transforming them into the frequency domain using the 3D Fourier Transform \(\mathcal{{F}_{\text{3D}}}\) \cite{freqdomain2}.
A learnable mask \(\mathbf{M}\) is applied to amplify frequency bands associated with vascular structures while suppressing irrelevant background frequencies. The enhanced features are then reconstructed using the inverse 3D Fourier Transform \(\mathcal{F}^{-1}_{\text{3D}}\) iresulting in a frequency token \(\mathbf{f}\) that captures global vascular structures by modeling low- and mid-frequency components. This process is mathematically expressed as:

\begin{equation}
\mathbf{f} = \mathcal{F}^{-1}_{\text{3D}}(\mathbf{M} \odot \mathcal{F}_{\text{3D}}(\mathbf{F}_{\mathrm{ASPP}})),
\label{FSA}
\end{equation}

where \(\odot\) denotes element-wise multiplication in the frequency domain.

The skip connection features (\(\mathbf{F}_{Skip}\)), edge token (\(\mathbf{E}\)), and frequency token (\(\mathbf{f}\)) are concatenated along the channel axis to form a composite tensor, \(\mathbf{C} \in \mathbb{R}^{B \times 3C \times D \times H \times W}\). This tensor is then processed using a 3D depthwise convolution \cite{dynamicdepthwiseconv}, which efficiently captures local spatial interactions while preserving the unique characteristics of each feature token. To further refine the composite features, an Efficient Channel Attention (ECA) module \cite{ECANET} is applied, adaptively recalibrating the importance of each channel and amplifying the most relevant features for vascular segmentation.

The final refined feature map (\(\hat{\mathbf{Y}}\)) is generated through a residual-style summation. A \(1 \times 1 \times 1\) convolution is applied to both the original ASPP feature map and the refined composite tensor, followed by summation to preserve foundational semantic features while integrating the enriched context from the edge and frequency tokens. The final output of the $MSC^2F$ module, \(\mathbf{F}_{MSC^2F}\)
, provides the decoder with a highly refined representation that combines multi-scale, edge-sensitive, and frequency-enhanced features, enabling precise segmentation of cerebrovascular structures.

\begin{equation}
\mathbf{F}_{\mathrm{MSC^2F}} = \text{Conv}_{1\times1\times1}(\mathbf{F}_{\mathrm{ASPP}}) + \text{Conv}_{1\times1\times1}(\hat{\mathbf{Y}}).
\end{equation}

\subsubsection{$CDA^2F$ Module}

The $CDA^2F$ module, illustrated in Fig.~\ref{fig:sefa-blocks}(b), is a core component of the NeuroVascU-Net architecture designed to integrate complementary feature extraction pathways and enhance hierarchical feature representation. This module is structured to process input features \(\mathbf{X} \in \mathbb{R}^{B \times C \times D \times H \times W}\) through four parallel branches, each tailored to address specific challenges posed by the intricate and variable morphology of the cerebrovascular network. The first branch utilizes 3D Involution \cite{involution} was used to generate location-specific, channel-shared kernels, allowing adaptive spatial filtering. The branch was employed to apply adaptive spatial filtering through location-specific, channel-shared kernels. This was critical for distinguishing fine vessels from surrounding tissue. Second, since contrast-enhanced vessels appear as high-frequency details in T1CE MRIs, this characteristic was leveraged by a Frequency Spatial Attention (FSA) module \cite{freqdomain2} which transforms input features into the frequency domain using a 3D Fourier Transform and applies a learnable spectral mask to selectively amplify vessel-specific frequency components while suppressing background noise. This process leverages the fact that contrast-enhanced vessels in T1CE MRI often manifest as high-frequency details.

To ensure robust feature extraction across all orientations of the vascular structures, the third branch integrates Spherical CNNs \cite{affineeq} to learn rotationally equivariant representations. This allows the module to accurately identify vascular patterns regardless of the anatomical orientation or imaging plane. Finally, the fourth branch utilizes a 3D Depthwise ConvNeXt block \cite{convnext} with large \(5 \times 5 \times 5\) kernels and GELU activation, which effectively captures both local and global contextual information while optimizing computational efficiency. 

The outputs from the Involution, FSA, and Spherical CNN branches are initially fused through summation, producing a composite tensor that is then scaled and merged with the output of the Depthwise ConvNeXt branch. This results in a rich multi-domain feature map \(\mathbf{F}_{\mathrm{MultiDomain}}\), which combines spatial, frequency, and geometric information. To further enhance the representation, a stochastic depth layer is applied, followed by a residual addition of the refined map to the initial input \(\mathbf{X}\) resulted in \(\mathbf{F}'_{\mathrm{MultiDomain}}\). This ensures that the module maintains the semantic richness of the input features while enhancing their geometric and spatial representations.

The final step of the this module involves processing the refined feature map through a gated axial transformer layer \cite{gatedffn}, which captures long-range dependencies and positional information by applying sequential attention along each spatial axis. A gating mechanism regulates the flow of information, selectively amplifying the most relevant features for vessel segmentation while suppressing less critical details. Through the integration of these specialized pathways, the $CDA^2F$ module generates robust and comprehensive feature representations that are vital for the segmentation of complex cerebrovascular structures. Positioned at both the encoder and decoder stages, the $CDA^2F$ module ensures consistent refinement of features across hierarchical levels, allowing the NeuroVascU-Net to achieve high segmentation accuracy in anatomically challenging regions.

\subsection{Evaluation Metrics}

The performance of the cerebrovascular segmentation models was quantitatively evaluated using a comprehensive suite of standard voxel-level metrics. These metrics were computed from the counts of true positives (TP), true negatives (TN), false positives (FP), and false negatives (FN) determined by comparing the model's predictions against the ground-truth (GT) annotations for the vessel class.
The primary overlap measures employed were the DSC, which effectively balances precision and recall and is particularly informative in scenarios with class imbalance:
\begin{equation}
\mathrm{DSC} = \frac{2 \cdot \mathrm{TP}}{2 \cdot \mathrm{TP} + \mathrm{FP} + \mathrm{FN}}
\end{equation}

and the Jaccard Index (JI), which quantifies the extent of overlap between the predicted segmentation and the ground truth, penalizing both over-segmentation and under-segmentation:

\begin{equation}
\mathrm{JI} = \frac{\mathrm{TP}}{\mathrm{TP} + \mathrm{FP} + \mathrm{FN}}
\end{equation}

Additional metrics were utilized to provide a more nuanced understanding of model performance. Sensitivity (Sens) was calculated as $(\mathrm{TP}/(\mathrm{TP} + \mathrm{FN}))$, indicating the model's ability to correctly identify true vessel voxels. Specificity (Spec) was determined as $(\mathrm{TN}/(\mathrm{TN} + \mathrm{FP}))$, evaluating the model’s proficiency in correctly recognizing background voxels, thereby minimizing false positive detections in non-vessel regions. Precision (Prec) was computed as $(\mathrm{TP}/(\mathrm{TP} + \mathrm{FP}))$, reflecting the proportion of voxels predicted as vessels that were genuine vessel structures.

Beyond segmentation accuracy, the computational efficiency and complexity of the models were also considered. This was assessed by reporting the total Number of Parameters, indicative of model size and memory requirements (Params), and the average Inference Time per volumetric scan, which reflects the model's speed during deployment.

\subsection{Hybrid Loss Function}

The pre-processed training dataset contained approximately 693 million voxels, exhibiting substantial class imbalance: 98.10\% background, 1.40\% vessel, and 0.50\% tumor voxels. This extreme imbalance, common in volumetric vessel segmentation tasks, can severely bias learning toward the dominant class.

To counteract this, a hybrid loss function was devised. This function integrates the advantages of Weighted Cross-Entropy (WCE) loss and Dice loss. The WCE component enabled differential penalization of misclassified voxels from under-represented classes, whereas the Dice component directly maximized spatial overlap between predicted and GT segmentations. The total loss function, denoted as \(\mathcal{L}_{\mathrm{Total}}\), was defined as a weighted sum of the two components:

\begin{equation}
    \mathcal{L}_{\mathrm{Total}} = \alpha\,\mathcal{L}_{\mathrm{WCE}} + \beta\,\mathcal{L}_{\mathrm{Dice}},
    \label{eq:total_loss}
\end{equation}

where \(\alpha\) and \(\beta\) represented hyperparameters that governed the relative contribution of each loss term.

The WCE loss, \(\mathcal{L}_{\mathrm{WCE}}\), was formulated to impose larger penalties on misclassifications of minority classes. Given a GT label \(y_i \in \{0,1\}\) and a predicted probability \(\hat{y}_i\) for class \(i\), the WCE loss was defined as:

\begin{equation}
    \mathcal{L}_{\mathrm{WCE}} = - \sum_{i=1}^{C} w_i \,y_i \,\log(\hat{y}_i),
    \label{eq:wce_loss}
\end{equation}

where \(w_i\) denoted the class-specific weight for class \(i\), and \(C\) represented the total number of classes.

To further enhance spatial coherence in segmentation outputs, the Dice loss component, \(\mathcal{L}_{\mathrm{Dice}}\), was employed. It was derived from the DSC, which quantifies overlap between the predicted segmentation and the GT mask. Given the flattened prediction vector \(p \in [0,1]^N\) and the binary GT vector \(t \in \{0,1\}^N\) across \(N\) voxels, the DSC was calculated as:

\begin{equation}
    \mathrm{DSC} = \frac{2 \sum_{j=1}^{N} p_j \,t_j + \varepsilon}{\sum_{j=1}^{N} p_j + \sum_{j=1}^{N} t_j + \varepsilon},
    \label{eq:dsc_coeff}
\end{equation}

where \(\varepsilon = 10^{-5}\) was used to prevent division by zero and enhance numerical stability. The Dice loss was then defined as:

\begin{equation}
    \mathcal{L}_{\mathrm{Dice}} = 1 - \mathrm{DSC}.
    \label{eq:dice_loss}
\end{equation}

\subsection{Optimization and Hyperparameter Configuration}

The NeuroVascU-Net was implemented using the PyTorch framework, leveraging the MONAI (Medical Open Network for AI) library \cite{cardoso2022monai} to facilitate efficient medical imaging workflows, including pre-processing and data augmentation. The dataset was partitioned into three subsets: 100 subjects were allocated for model training, 10 subjects were used for validation to tune hyperparameters and monitor training progress, and the remaining 27 subjects were reserved as an independent test set for assessing the model’s generalization capabilities. This division ensured robust evaluation of the model’s performance across unseen data and minimized the risk of overfitting.

The Adam optimizer was employed for training, selected for its adaptive learning rate capabilities and proven effectiveness in stabilizing convergence in DL tasks. A learning rate of \(8 \times 10^{-5}\) was chosen after systematic experimentation, striking a balance between rapid convergence and stable training dynamics. To mitigate overfitting, an early stopping mechanism was integrated into the training pipeline, monitoring validation loss and halting training when performance plateaued. This strategy ensured optimal model performance without excessive training that could lead to overfitting.

The loss function was a hybrid formulation combining Weighted Cross-Entropy (WCE) loss and Dice loss, specifically tailored for the challenges of cerebrovascular segmentation in T1CE MRI. The WCE loss incorporated a class weight of 8.546 for the vessel class, calculated as the square root of the background-to-vessel voxel ratio (73.52). This weighting addressed the significant class imbalance by amplifying the contribution of vessel-related voxels during training. The hybrid loss coefficients were fine-tuned to $\alpha$ = 2.0 for WCE and $\beta$ = 1.0
 for Dice loss, as expressed in Eq.~\eqref{eq:total_loss}. This configuration demonstrated superior segmentation performance during validation, effectively balancing pixel-wise classification accuracy with global overlap metrics. Alternative loss functions, such as focal loss, were explored during preliminary experiments but did not achieve comparable results.

Training was conducted on an NVIDIA Tesla P100 GPU, which provided sufficient computational power for handling high-resolution 3D medical imaging data. To further reduce overfitting, a dropout rate of 0.2 was applied within the network, randomly dropping connections during training to improve generalization. A batch size of 2 was chosen to optimize GPU memory usage while maintaining stable convergence. This smaller batch size allowed the model to process high-detail data without exceeding hardware constraints.

\section{Experimental Results}

\begin{figure}
    \centering
    \begin{subfigure}{0.49\linewidth}
        \centering
        \includegraphics[width=\linewidth]{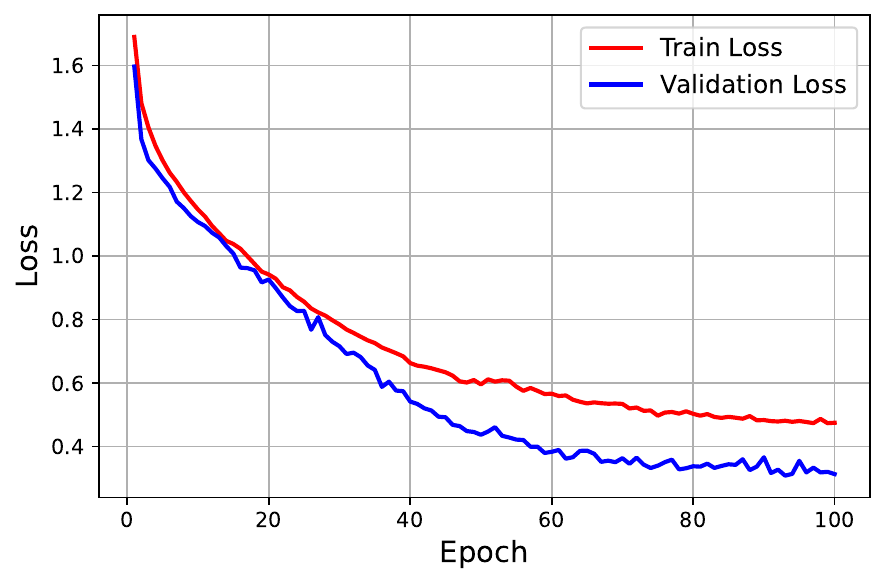}
        \caption{}
    \end{subfigure}
    \begin{subfigure}{0.49\linewidth}
        \centering
        \includegraphics[width=\linewidth]{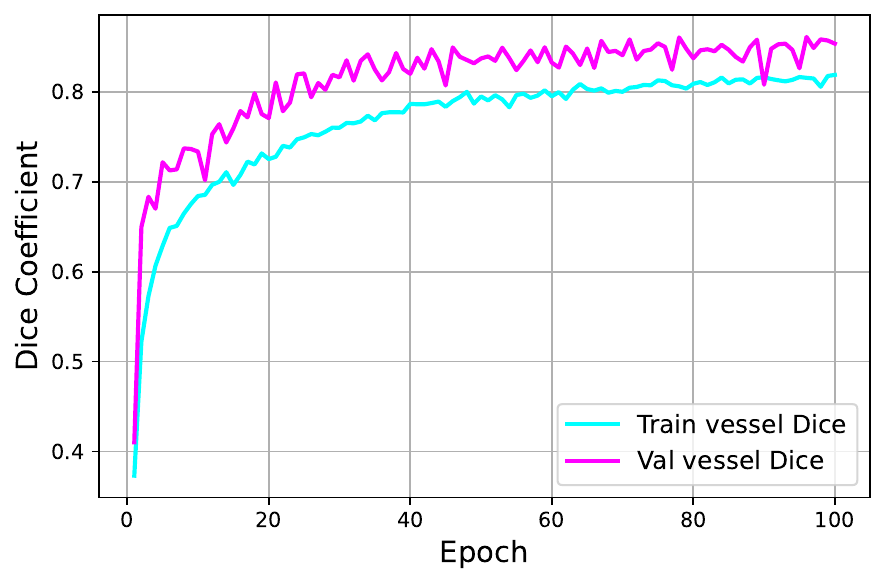}
        \caption{}
    \end{subfigure}
    \caption{NeuroVascU-Net performance over epochs (a) Train and overall validation loss. (b) Train and overall validation DSC, the lower training DSC in comparison to validation is caused by sliding window inference for validation data.}
    \label{fig:plots}
\end{figure}

The proposed NeuroVascU-Net architecture exhibited robust performance and high-fidelity segmentation of cerebrovascular structures on the in-house T1CE MRI dataset. The progression of training and validation results is illustrated in Fig.~\ref{fig:plots}, where panel (a) demonstrates stable convergence of the training and validation loss curves, and panel (b) highlights the steady improvement of the DSC across epochs. Notably, the lower training DSC compared to validation DSC is attributed to the sliding window inference strategy used for validation data, ensuring unbiased evaluation during training.

Quantitative metrics, summarized in Table~\ref{tab:performance_comparison}, underscore the model’s state-of-the-art performance. NeuroVascU-Net achieved a mean DSC of 0.8609 and a Jaccard Index (JI) of 0.7582, reflecting excellent spatial overlap with ground-truth annotations. The model demonstrated a high precision of 0.8841, indicating its ability to accurately delineate vessel boundaries with minimal false positives. Additionally, the specificity of 0.9982 highlights the model’s strong capability to exclude irrelevant background regions, further improving segmentation reliability. Although the sensitivity of 0.8456 was slightly lower compared to precision, it still demonstrated the model’s robust ability to identify true vessel voxels in highly complex anatomical regions.

In terms of computational efficiency, the NeuroVascU-Net architecture comprises 12.43 million trainable parameters and achieved an average inference time of 3,840 seconds per volume, which is competitive given the model’s complexity and the high-resolution input data. Compared to other benchmark models (Table~\ref{tab:performance_comparison}), NeuroVascU-Net delivered superior DSC, Jaccard Index, precision, and specificity, while maintaining a reasonable computational footprint.

To validate the contributions of the NeuroVascU-Net’s specialized modules, extensive ablation studies were conducted. Removal of the $MSC^2F$ module from the bottleneck led to a significant decline in segmentation performance, confirming its critical role in capturing multi-scale spatial context, edge-boundary details, and frequency-domain information. Similarly, replacing the $CDA^2F$ module at the fourth hierarchical level with standard dilated convolution blocks resulted in reduced segmentation accuracy. This highlights the importance of the $CDA^2F$ module’s ability to learn complex, abstract representations through its integration of geometric invariance, spatial filtering, and frequency-based attention mechanisms. These experiments affirm that both modules are integral to achieving the model’s high accuracy and generalization performance.

\begin{table}
\centering
\caption{Performance comparison for brain vessel segmentation on T1CE dataset trained from scratch.}
\label{tab:performance_comparison}
\resizebox{1\columnwidth}{!}{%
  \begin{tabular}{lccccccc}
    \toprule
    \textbf{Model} &
      \textbf{DSC} &
      \textbf{JI} &
      \textbf{Sens} &
      \textbf{Spec} &
      \textbf{Prec} &
      \textbf{Params} &
      \makecell{\textbf{Inf. Time}\\\textbf{Sec.}} \\
    \midrule
    U-Net \cite{Unet3Dvessel}            & 0.7220          & 0.5801          & 0.8762          & 0.9901          & 0.5780          & 19,283,541 & \textbf{1,250} \\ 
    Attention U-Net \cite{attentionunet} & 0.8023          & 0.6723          & \textbf{0.8835} & 0.9951          & 0.7411          & 23,674,382 & 1,450          \\ 
    MDA U-Net (3D impl.) \cite{mdaunet}  & 0.8062          & 0.6781          & 0.8766          & 0.9955          & 0.7598          & 12,709,318 & 4,240          \\ 
    U-Net +CABM+MSFE \cite{MSFE}         & 0.8100          & 0.6839          & 0.8182          & 0.9971          & 0.8342          & 9,884,354  & 3,990          \\ 
    Brave-Net \cite{Bravenet}            & 0.8308          & 0.7131          & 0.8423          & 0.9973          & 0.8258          & 10,911,813 & 4,400          \\ 
    Dilated attention U-Net+ ASPP         & 0.8424          & 0.7303          & 0.8688          & 0.9971          & 0.8279          & 13,407,190 & 2,240          \\ 
    Swin U-NetR \cite{swinunetrvessel}   & 0.8600          & 0.7577          & 0.8814          & 0.9974          & 0.8454          & 15,703,004 & 4,480          \\ 
    The Proposed Model                  & \textbf{0.8609} & \textbf{0.7582} & 0.8456          & \textbf{0.9982} & \textbf{0.8841} & 12,429,814 & 3,840          \\ 
    \bottomrule
  \end{tabular}%
}
\end{table}

\begin{figure}[ht]
    \includegraphics[width=0.9\linewidth]{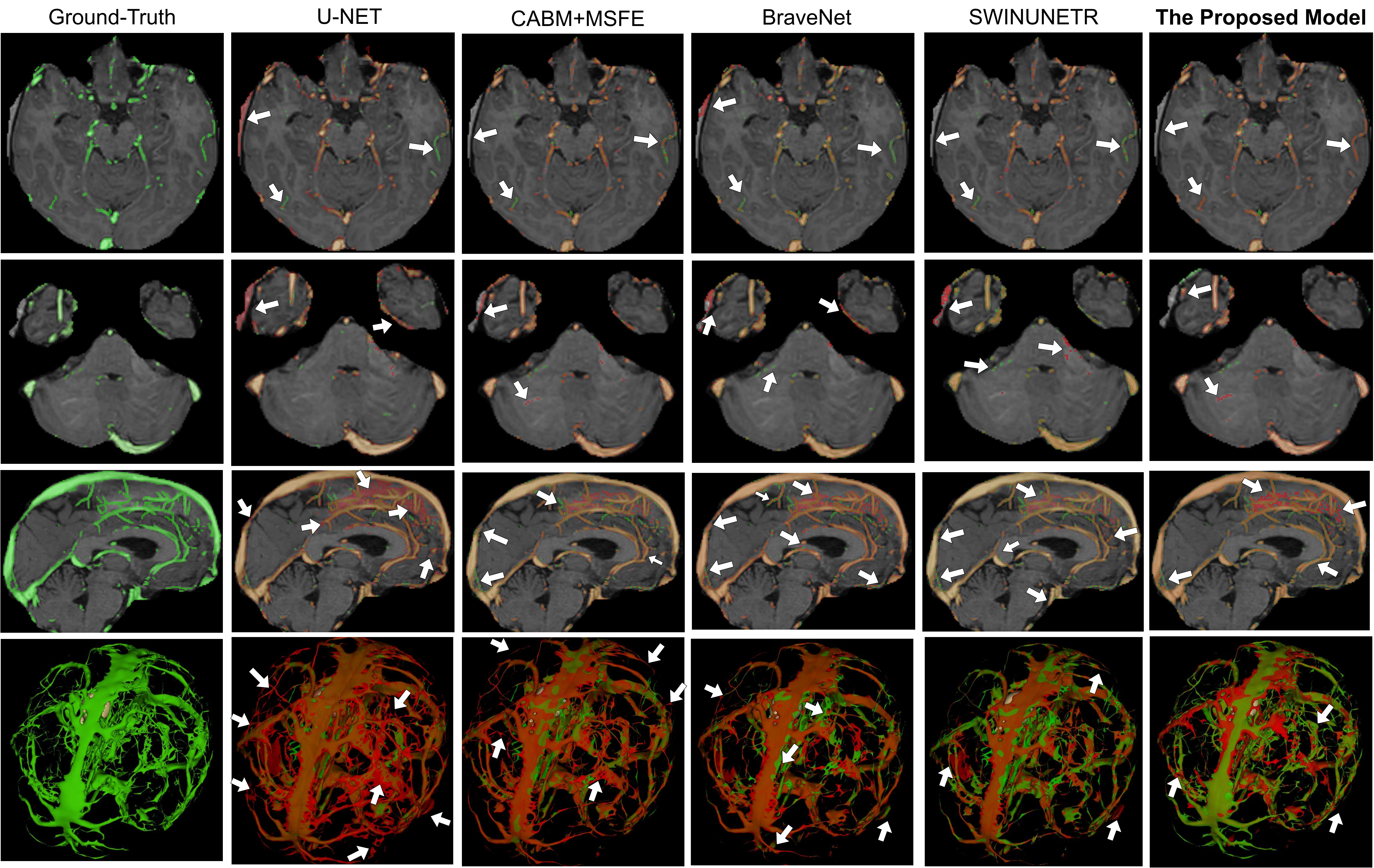}
    \caption{Visualization of brain vessel segmentation performance for each DL model on the T1CE dataset. The first column shows the ground truth (GT), and the following columns display model predictions. Rows 1–3 show sample prediction vs. GT slices from validation data, while the fourth row presents a 3D overlay with opacity. True Positives (TP) in orange, False Positives (FP) in red, and False Negatives (FN) in green are highlighted in each image.}
    \label{fig:3Dcompare}
\end{figure}

The qualitative results further corroborate the quantitative findings, as illustrated in Fig~\ref{fig:3Dcompare}. The NeuroVascU-Net demonstrated exceptional capability in identifying and delineating extremely fine, distal vessels, which are crucial for comprehensive surgical planning. The model was particularly adept at capturing intricate vascular networks surrounding tumors, showcasing its potential for direct application in neurosurgical workflows. While occasional false positives were observed in regions of high-intensity white matter, and minor undersegmentation of larger vessels occurred in certain orientations, the overall connectivity and detail of the segmented vascular tree were exceptionally accurate.

Visualization of segmentation outputs (Fig~\ref{fig:3Dcompare}) highlights the model’s ability to differentiate between true positives (TP, orange), false positives (FP, red), and false negatives (FN, green). Compared to alternative approaches such as U-Net \cite{Unet3Dvessel}, Attention U-Net \cite{attentionunet}, and Swin U-NetR \cite{swinunetrvessel}, NeuroVascU-Net consistently produced outputs that were closest to ground-truth annotations. The 3D reconstructions further emphasize the model’s superior ability to preserve vascular connectivity and delineate fine vessels, even in anatomically complex regions around tumors.

\section{Discussion}

This study introduces NeuroVascU-Net, a specialized deep learning architecture developed to address the significant challenge of 3D cerebrovascular segmentation from clinically standard T1CE MRI. The results demonstrate that the proposed model achieves state-of-the-art segmentation accuracy while maintaining computational efficiency, validating the core design philosophy that hybrid CNN architectures, enriched with task-specific modules, can outperform larger, generalized transformer-based architectures in both segmentation quality and practical applicability for clinical workflows.

A detailed comparative analysis, presented in Table~\ref{tab:performance_comparison}, positions NeuroVascU-Net as a leading solution among existing methods. Traditional benchmarks, such as standard U-Net~\cite{Unet3Dvessel} and Attention U-Net~\cite{attentionunet}, lack the specialized capabilities required to handle the nuances of T1CE imaging, particularly in precision and vessel delineation. Among the comparisons, the most notable is with Swin U-NetR~\cite{swinunetrvessel}, a transformer-based architecture. NeuroVascU-Net surpasses Swin U-NetR with a slightly higher Dice Similarity Coefficient (DSC) of 0.8609 vs. 0.8600 and, more importantly, a significantly higher Precision (0.8841 vs. 0.8454). The improved precision indicates fewer false positives and highlights NeuroVascU-Net’s ability to accurately delineate vessel boundaries. This advantage is largely attributable to the explicit edge-detection mechanism embedded within the $MSC^2F$  module, particularly the LoG filter, which is adept at capturing sharp contrast boundaries characteristic of vessels in T1CE MRI. In contrast, transformer models rely on patch-based mechanisms that may struggle to capture these structural priors as effectively.

Beyond its segmentation accuracy, NeuroVascU-Net demonstrates exceptional computational efficiency, operating with approximately 21\% fewer parameters than Swin U-NetR (12.4M vs. 15.7M). This reduction in model complexity directly benefits clinical deployment by lowering memory requirements and enabling faster inference times. Such efficiency is critical in hospital environments, where access to high-performance GPUs may be limited, and rapid processing is essential for pre-operative planning. By providing accurate segmentation with reduced computational overhead, NeuroVascU-Net emerges as a pragmatic and accessible solution tailored for real-world applications.

The architectural innovations embedded within NeuroVascU-Net are central to its success. Ablation studies confirmed that every component of the model is strategically designed and contributes uniquely to its performance. For instance, the placement of the $CDA^2F$ module at the fourth hierarchical level of the encoder enables the model to process highly abstract feature maps, distinguishing between geometrically complex structures such as tortuous vessels and enhancing tumor rims. The $CDA^2F$ module incorporates advanced mechanisms like adaptive spatial filtering (via 3D Involution) and rotational equivariance (via Spherical CNNs), which are critical for disentangling complex features at deeper network stages. Similarly, the $MSC^2F$ module positioned at the bottleneck enriches the compressed feature representation with multi-scale spatial context, edge-boundary sensitivity, and spectral information. This ensures that critical details are preserved and provides the decoder with refined input for segmentation reconstruction. For neurosurgeons, this translates to greater confidence in the segmentation maps, particularly when delineating safe surgical corridors near critical vascular structures.

This study also contributes to the ongoing discourse in medical imaging regarding architectural design principles. While transformer models have proven their versatility as general-purpose vision architectures, the results of NeuroVascU-Net highlight the strength of task-specific designs for problems with strong domain priors. For challenges such as cerebrovascular segmentation, where the tube-like geometry and sharp contrast boundaries of vessels are well-defined structural characteristics, explicitly embedding inductive biases can yield more efficient and accurate solutions than generalized complexity. NeuroVascU-Net exemplifies the principle that architectural specialization tailored to specific clinical tasks can outperform generalized models in both precision and practicality.

Despite its strengths, NeuroVascU-Net is not without limitations, which suggest promising avenues for future research. First, the model was validated on a single-center dataset, raising the possibility of biases related to specific imaging protocols and scanner characteristics. To ensure broader applicability, future studies should focus on evaluating NeuroVascU-Net’s performance across multi-institutional datasets with diverse imaging settings. Second, the segmentation of sub-voxel scale vessels remains an unresolved challenge due to the partial volume effect inherent in MRI. Addressing this frontier may require integrating super-resolution techniques or designing novel loss functions that account for the ambiguity of low-signal structures. Lastly, while NeuroVascU-Net is computationally efficient, further optimization techniques such as model pruning or quantization could be explored to enable near real-time performance for intraoperative applications, where rapid segmentation is critical.

\section{Conclusion}

This study introduces NeuroVascU-Net, a novel 3D DL architecture designed for automated segmentation of cerebrovascular structures from clinically standard T1-weighted post-contrast MRI scans. The work makes two key contributions: the curation and expert annotation of a specialized T1CE MRI dataset comprising 137 neuro-oncology patients, and the development of the NeuroVascU-Net architecture, which integrates advanced feature extraction mechanisms tailored to the complexities of cerebrovascular segmentation. The architecture incorporates a $MSC^2F$ module at the bottleneck and a $CDA^2F$ module at deeper hierarchical levels. These modules enable NeuroVascU-Net to effectively capture local, global, and domain-specific features, ensuring detailed vascular mapping across diverse scales and orientations. As a result, NeuroVascU-Net achieved state-of-the-art segmentation performance, with a DSC of 0.8609 and a precision of 0.8841, demonstrating its ability to accurately delineate both large and fine vascular structures. Notably, this high accuracy was achieved with superior computational efficiency, making NeuroVascU-Net a practical and accessible solution for clinical settings where computational resources may be limited. The clinical relevance of NeuroVascU-Net is underscored by its ability to generate precise, automated segmentations from T1CE MRI—the imaging modality most relevant to neuro-oncological surgery. By providing accurate vascular maps, NeuroVascU-Net has the potential to significantly enhance the safety and efficacy of neurosurgical procedures, particularly in the planning of surgical approaches and biopsies. Its integration into modern computer-assisted operating rooms positions NeuroVascU-Net as a valuable tool for improving patient outcomes and advancing neurosurgical workflows. Looking ahead, several promising avenues for future research remain. Expanding the segmentation capabilities of NeuroVascU-Net to include additional brain tissues, such as cerebrospinal fluid, white matter, and other anatomical structures, could further broaden its clinical utility. Additionally, the development of an intelligent system to optimize biopsy trajectories based on spatial relationships between tumors and critical vascular and tissue structures represents an important next step. Such advancements could enable safer and more effective neurosurgical planning by providing surgeons with detailed, actionable insights into the spatial organization of brain anatomy.

\bibliographystyle{elsarticle-num} 
\bibliography{cas-refs.bib}

\end{document}